\documentclass[final]{cvpr}

\usepackage{times}
\usepackage{epsfig}
\usepackage{graphicx}
\usepackage{amsmath}
\usepackage{amssymb}

\usepackage[pagebackref=true,breaklinks=true,colorlinks,bookmarks=false]{hyperref}

\def\cvprPaperID{3375} 
\def\confYear{CVPR 2021}

\begin{document}

\title{KShapeNet: Riemannian network on Kendall shape space for Skeleton based Action Recognition}


\author{Racha Friji\\
CRISTAL Lab, ENSI, La Manouba University, Tunisia\\
Talan Innovation Factory, Talan\\
{\tt\small racha.friji@talan.com}
\and
Hassen Drira\\
IMT Lille Douai, Institut Mines-Télécom,\\ Centre for Digital Systems, F-59000 Lille, France\\
Univ. Lille, CNRS, Centrale Lille, Institut Mines-Télécom, \\UMR 9189 - CRIStAL - F-59000 Lille, France\\
\and
Faten Chaieb\\
AllianSTIC Laboratory, \\ EFREI PARIS, France\\
CRISTAL Lab, ENSI, \\ La Manouba University, Tunisia\\
\and
Sebastian Kurtek\\
Department of Statistics,\\ The Ohio State University, USA
\and
Hamza Kchok,\\ Ecole Nationale des Sciences de l’Informatique \\INSAT, Tunisia
}

\maketitle

\begin{abstract}


Deep Learning architectures, albeit successful in most computer vision tasks, were designed for data with an underlying Euclidean structure, which is not usually fulfilled since pre-processed data may lie on a non-linear space.
In this paper, we propose a geometry aware deep learning approach for skeleton-based action recognition. 
Skeleton sequences are first modeled as trajectories on Kendall’s shape space and then mapped to the linear tangent space. The resulting structured data are then fed to a deep learning architecture, which includes a layer that optimizes over rigid and non rigid transformations of the 3D skeletons, followed by a CNN-LSTM network. The assessment on two large scale skeleton datasets, namely NTU-RGB+D and NTU-RGB+D 120, has proven that proposed approach outperforms existing geometric deep learning methods and is competitive with respect to recently published approaches. 



\end{abstract}

\section{Introduction}


Human behavior analysis via diverse data types has emerged as an active research issue in computer vision due to 1) the wide spectrum of not yet fully explored application domains, e.g., human-computer interaction, intelligent surveillance security, virtual reality, etc., and 2) the development of advanced sensors such as Intel RealSense, Asus Xtion and the Microsoft Kinect\cite{Author1}, which yield various data modalities, e.g., RGB and depth image sequences, and videos. Conventionally, these modalities have been utilized, solely \cite{Author4, Author5}, or merged (e.g., RGB + optical flow), for action recognition tasks \cite{Author8,Author9} using multiple classification techniques, and resulted in excellent results. With the development of human pose estimation algorithms \cite{Author2,Author3}, the problem of human joints (i.e., key-points) localization was solved and the acquisition of accurate 3D skeleton data reliably became possible. In comparison with former modalities, skeleton data, a topological representation of the human body using joints and bones, appears to be less computationally expensive, more robust in front of intricate backgrounds and variable conditions in terms of the viewpoint, scales and motion speeds.
An efficient way to analyze 3D skeleton motions is to consider their shapes independently of undesirable transformations; the resulting representation space of skeleton data is then non linear. 

Accordingly, we represent 3D skeleton landmarks in the Kendall shape space \cite{kendall1984shape} that defines shape as the geometric information that remains when location, scaling and rotational
effects are filtered out. A sequence of skeletons is then modeled as a trajectory on this space.
Thus, to analyze and classify such data, it is more suitable to consider the geometry of the underlying space. This remains a challenging problem since most commonly used techniques were designed for linear data.
Deep learning architectures, despite their efficiency in many computer vision applications, usually ignore the geometry of the underlying data space. Therefore, geometric deep learning architectures have been introduced to remedy this issue.
To the best of our knowledge, the main previous geometric deep learning approaches were designed on feature spaces (e.g., SPD matrices, Grassmann manifold, Lie groups \cite{Author44,Author41}) or on the 3D human body manifold \cite{Bronstein1,Bronstein2}. The literature that considers this problem on shape spaces is scarce. Actually, an extension of a conventional deep architecture on a pre-shape space has been recently proposed in \cite{Author40}, and an auto encoder-decoder has been extended to a shape space for gait analysis in \cite{Author48}.

In this work, we propose a novel geometric deep learning approach on Kendall's shape space, denoted KShapeNet, for skeleton-based action recognition. 
Skeleton sequences are first modeled as trajectories on Kendall’s shape space by filtering out scale and rigid transformations. Then, the sequences are mapped to a linear tangent space and the resulting structured data are fed into a deep learning architecture. The latter includes a novel layer that learns the best rigid or non rigid transformation to be applied to the 3D skeletons to accurately recognize the actions.
\paragraph{Contributions}
The main contributions of this paper are:
\begin{enumerate}
    \item We introduce a novel deep architecture on Kendall's shape space that deeply learns transformations of the skeletons for action recognition tasks.
    \item The proposed deep network includes a novel transformation layer that optimizes over rigid and non rigid transformations of skeletons to increase action recognition accuracy.
    \item 
    Experiments are conducted on two large scale publicly available datasets, NTU-RGB+D and NTU-RGB+D120, to show the competitiveness of the proposed approach in the context of 3D action recognition.
    
\end{enumerate}

\paragraph{Organization of the paper}
The rest of the paper is organized as follows. In Section \ref{section:Related work},
we briefly review existing solutions for action recognition and geometric deep learning. Section \ref{section:Modeling of shape space trajectories} describes geometric modeling of skeleton trajectories on Kendall's shape space. In Section \ref{section:TransformationLayer}, we introduce the proposed geometric deep architecture, KShapeNet. Experimental settings, results and discussions are reported in Section \ref{section:experiments}. Section \ref{section:conclusion} concludes the paper and summarizes a few directions for future work.
\section{Related work}
\label{section:Related work}
Geometric deep learning for action recognition has recently attracted a lot of attention from the research community. This has resulted in a variety of related approaches. Accordingly, we focus on highlighting the main categories in the areas of 3D action recognition and geometric deep learning. Interested readers can find exhaustive details in the associated recent surveys \cite{Author12, Author13}.

\subsection{Action recognition}

Presently, deep learning methods for human action recognition are preferred over traditional skeleton-based ones, which tend to focus on extracting hand crafted features \cite{Author14,Author15} The former methods can be categorized into three major sets: methods based on Recurrent Neural Network (RNN) \cite{Author16}, methods based on Convolutional Neural Network (CNN) \cite{Author17}, and methods based on Graph Convolutional Network (GCN) \cite{Author18}.

Since RNNs are convenient for time series data processing, RNN-based methods consider skeleton sequences as time series of coordinates of the joints. For the purpose of improving the  capability of learning the temporal context of skeleton sequences, Long Short Term Memory (LSTM) and Gated Recurrent Unit (GRU) have been introduced as efficient alternatives for skeleton based action recognition. Zhu et al. \cite{Author26} used an LSTM network and characterized joints through  the co-occurrence between actions. In \cite{Author27}, geometric joint features are applied to a multi layered LSTM network instead of passing directly passing in the joint positions. 
The pitfall of some of these methods \cite{Author20,Author21} is their weak ability of spatial modeling, resulting in non competitive results. A novel two-stream RNN architecture was recently proposed by Hong and Liang \cite{Author22}. This architecture models both the temporal dynamics and spatial configurations of skeleton data by applying an exchange of the skeleton axes at data level pre-processing. Relatedly, Jun and Amir \cite{Author29} focused on extracting the hidden relationship between the two domains (spatial and temporal) using a traversal approach on a given skeleton sequence. Unlike the general  method where joints are arranged in a simple chain ignoring kinectic dependency relations between adjacent joints, this tree-structure based traversal does not add false connections between body joints when their relation is not strong enough. Another pitfall of RNN based methods are the gradient exploding and vanishing problems over layers. Some new RNN architectures \cite{Author23,Author24} were proposed to address this particular limitation.

CNN models have excellent capability to extract high level  information and semantic cues. Multiple works \cite{ Author30, Author31,Author33} have exploited CNN models for action recognition by encoding the skeleton joints as images or pseudo-images prior to feeding them to the network. In \cite{Author33}, Zhang et al. map a skeleton sequence  to an image, referred to as the skeleton map, to facilitate spatio temporal dynamics modeling via the ConvNet. The challenge with CNN based methods is the extraction and utilization of spatial as well as temporal information from 3D skeleton sequences. Several other problems hinder these techniques including model size and speed  \cite{Author34}, occlusions, CNN architecture definition \cite{Author35}, and viewpoint variations \cite{Author33}. Skeleton based action recognition using CNNs thus remains a not completely solved research question.

Recently, the GCN has been
adopted to action recognition. This network represents human 3D skeleton data as a graph. There are two main types of
graph related neural networks: the graph recurrent neural network, and the graph
convolutional neural network \cite{Author36,Author37}.

\begin{figure*}[!t]
\begin{center}
   \includegraphics[width=\linewidth]{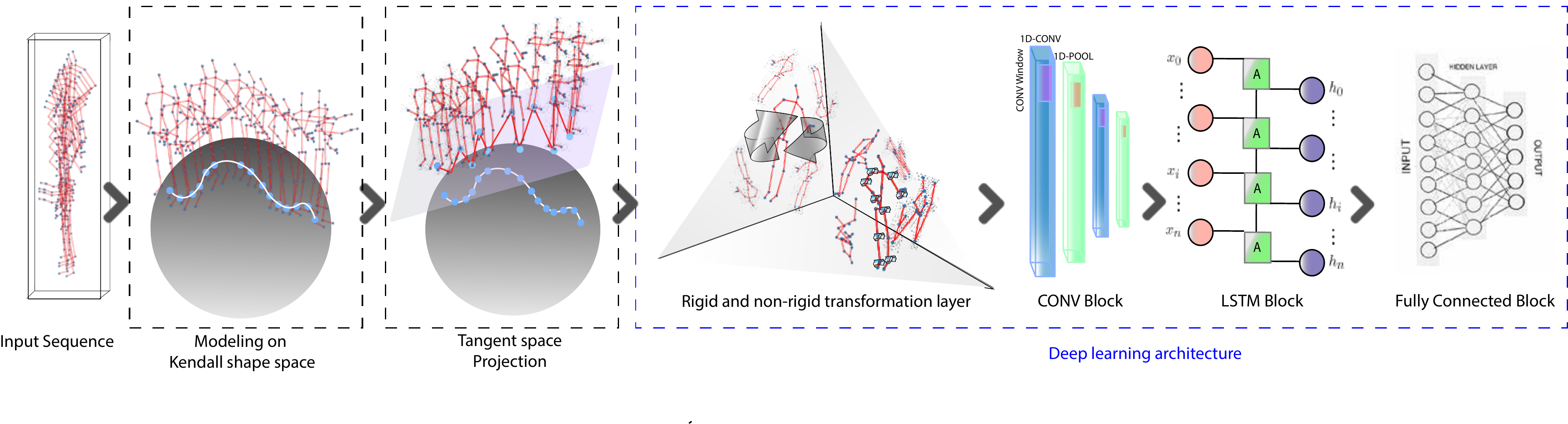}
\end{center}
   \caption{Illustration of the full architecture of the proposed KShapeNet, and its different blocks.}
\label{KShapeNet}
\end{figure*}

\subsection{Geometric deep learning}

Compared to previous techniques, geometric deep learning is a nascent research area. As mentioned earlier, it studies the extension of existing deep learning frameworks and algorithms to effectively process graph and manifold data. 
Some manifold based techniques have proven their success in 3D human action recognition due to view invariance of the manifold based representation of skeletal data. As examples, we cite the projection on Riemanian manifold \cite{Author40}, shape silhouettes in Kendall’s shape space \cite{Author38}, and linear dynamical systems on the Grassmann manifold
\cite{Author39}. Geometric deep learning approaches can be categorized into two main classes: approaches on manifolds and approaches on graphs. This paper is related to deep approaches on manifolds, and thus, we give a quick review of the state-of-the-art in this category.

Manifold-based geometric deep learning approaches extend deep architectures to Riemannian manifolds, interpreted either as feature spaces  \cite{Author44,Author45,Author41} or the human body shape (i.e., the human body is viewed as a manifold) \cite{Bronstein1,Bronstein2}.
Huang et al. proposed several networks on non linear manifolds. In \cite{Author44}, they introduced the first network architecture to perform deep learning on the Grassmann manifold. They presented competitive results on three datasets of emotion recognition, action recognition and face verification, respectively. Along similar lines, an architecture on the manifold of SPD matrices was proposed in \cite{Author45}, and similar experimental evaluation proved the effectiveness of this approach. Recently, the same authors proposed an architecture on Lie groups with application to skeleton-based action recognition \cite{Author41}. These approaches investigate the non-linearity of various feature spaces, but did not consider shape spaces. Limited efforts have recently been made to design deep architectures on some shape-preshape spaces. Friji et al. \cite{Author40} proposed a deep architecture on the sphere for modeling unit-norm skeletons with application to action recognition. Along similar lines, Hosni et al. \cite{Author48} extended the auto-encoder to a shape space with application to gait recognition.


\section{Modeling of shape space trajectories}
\label{section:Modeling of shape space trajectories}

We use the landmark shape based representation of the human skeleton, and geometric tools from Kendall’s shape analysis \cite{Author60} to model skeleton shapes and their temporal evolution. Every point in the shape space represents a single static action shape, and the distance between two such points illustrates the magnitude of shape discrepancies between the respective shapes. 



Each skeleton $X$ in an action sequence is represented as a set of $n$ landmarks in $\mathbb{R}^3$, i.e., $X \in \mathbb{R}^{n\times 3}$. In our framework, we model skeletal shape sequences and use Kendall's shape representation to achieve the required invariances with respect to translation, scale and rotation. First, we perform data interpolation via cubic splines, to have the same number of frames for each sequence, rather than the commonly used zero-padding technique. 

Translation and scale variabilities can be removed from the representation space via normalization as follows. Let $H$ denote the $(n-1) \times n$ sub-matrix of a Helmert matrix, as detailed in \cite{Author60}, where the first row is removed. In order to center a skeleton $X$, we pre-multiply it by $H$, $HX \in \mathbb{R}^{(n-1)\times 3}$; then, $HX$ contains the centered Euclidean coordinates of $X$. Let $C_0=\{HX \in \mathbb{R}^{(n-1)\times 3} 
|X \in \mathbb{R}^{n\times 3}\}$, which is a $3(n-1)$ dimensional vector space, which can be identified with  $\mathbb{R}^{3(n-1)}$. Using the standard Euclidean inner product (norm) on $C_0$, we scale all centered skeletons to have unit norm. As a result, we define the pre-shape space as $C = \{HX \in C_0| \|HX\|^2 = (HX)^T(HX) = 1\}$; due to the unit norm constraint, $C$ is a $(3n-4)$-dimensional unit sphere in $\mathbb{R}^{3(n-1)}$. Henceforth, we will refer to an element of $C$ as $\tilde{X}$, i.e., a centered and unit norm skeleton. The tangent space at any pre-shape $\tilde X$ is given by $T_{\tilde{X}}(C) = \{V \in \mathbb{R}^{3n-4}| \langle V,X\rangle = V^TX=0\}$. 

\begin{figure}[ht]
    \centering
    \includegraphics[ width=\linewidth]{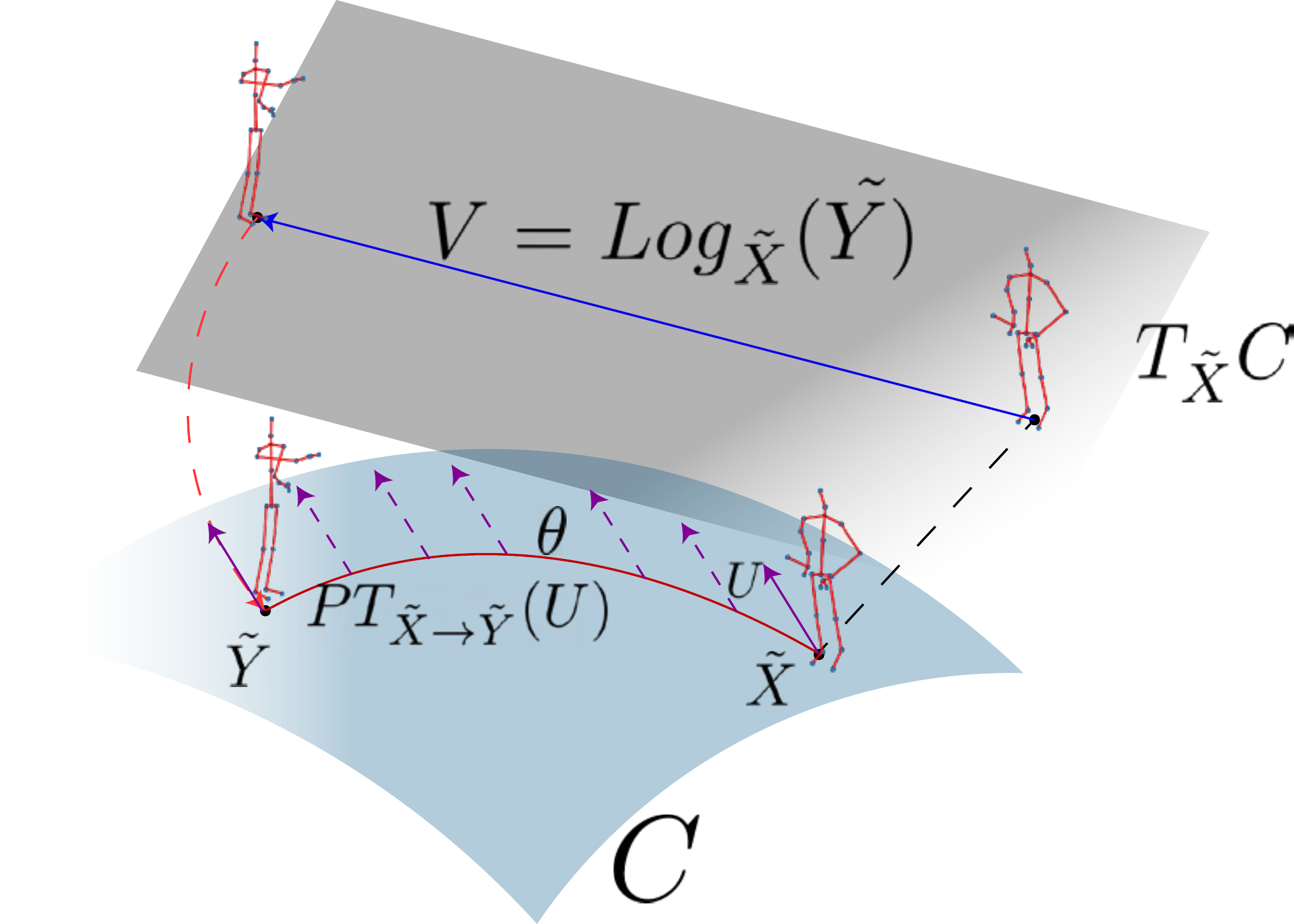}
    \caption{Illustration of the logarithmic mapping and parallel transport techniques on Kendall's pre-shape space.}
    \label{PT}
\end{figure}

In subsequent analyses, our representation of skeleton sequences further passes to the tangent space. Thus, it is useful to define three Riemannian geometric tools that allow one to map points 1) from the pre-shape space to a tangent space, 2) from a tangent space to the pre-shape space, and 3) between different tangent spaces. Task 1) can be achieved via the logarithmic map, $log_{\tilde{X}}: C \to T_{\tilde X}(C)$, defined as (for $\tilde{X},\ \tilde{Y}\in C$):
\begin{equation}
\label{eq2}
 \centering
log_{\tilde{X}}(\tilde{Y}) = \frac{\theta}{\sin({\theta})} (\tilde{Y} - \cos(\theta)\tilde{X}),
\end{equation}
where $\theta=\cos^{-1}\left(\langle\tilde{X},\tilde{Y}\rangle\right)$ is the arc-length distance between $\tilde{X}$ and $\tilde{Y}$ on $C$. Task 2) is carried out via the exponential map, $exp_{\tilde{X}}: T_{\tilde{X}}(C)\to C$, defined as (for $\tilde{X}\in C$ and $V \in T_{\tilde{X}}(C)$):
\begin{equation}
\label{eq1}
 \centering
\tilde{Y}= \cos(\|V\|)\tilde{X}+\sin(\|V\|)\frac{V}{\|V\|},
\end{equation}
where $\|V\|=\sqrt{V^TV}$ as before. Finally, for task 3), we use parallel transport, which, in short, defines an isometric mapping between tangent spaces. The parallel transport, $PT_{\tilde{X}\to \tilde{Y}}:T_{\tilde{X}}(C)\to T_{\tilde{Y}}(C)$ is defined as (for $\tilde{X},\ \tilde{Y}\in C$ and $U\in T_{\tilde{X}}(C)$): 
\begin{equation}
\label{eq3}
PT_{\tilde{X}\to\tilde{Y}}(U)=
U - \frac{\langle log_{\tilde{X}}(\tilde{Y}),U\rangle}{\theta}
\left(log_{\tilde{Y}}(\tilde{X}) + log_{\tilde{X}}(\tilde{Y})\right),
\end{equation}
where $\langle\cdot,\cdot\rangle$ and $\theta$ are the standard Euclidean inner product and the distance between $\tilde{X}$ and $\tilde{Y}$ on $C$, respectively, as before. A pictorial description of the logarithmic and exponential maps, and parallel transport, is given in Fig. \ref{PT}.
While translation and scale can be dealt with through normalization, rotation variability in Kendall's framework is removed algebraically using the notion of equivalence classes. The rotation group in $\mathbb{R}^3$ is given by $SO(3)=\{O\in \mathbb{R}^{3\times 3}|O^TO=I,\ det(O)=1\}$. For $O\in SO(3)$ and $\tilde{X}\in C$, the action of the rotation group is given by matrix multiplication, i.e., $O\tilde{X}$ is a rotation of $\tilde{X}$. Let $[\tilde{X}]=\{O\tilde{X}|O\in SO(3),\ \tilde{X}\in C\}$ denote an equivalence class of a pre-shape $\tilde{X}$. Then, Kendall's shape space is the quotient space $C/SO(3)$. Rotation variability is removed in a pairwise manner (or with respect to a given template), by optimally aligning two configurations $\tilde{X}$ and $\tilde{Y}$ via Procrustes analysis \cite{Author60}; we omit the details of this process here for brevity. After optimal rotation, one can use the same Riemannian geometric tools as on the pre-shape space $C$, e.g., Equations \ref{eq2}-\ref{eq3}, to model shapes of skeleton landmark configurations.

\section{Shape space deep architecture} 
\label{section:TransformationLayer}
The proposed deep learning architecture of Kendall's Shape Space Network, KShapeNet, is illustrated in Fig. \ref{KShapeNet}. 
Input skeleton sequences are first modeled as trajectories on $C$, after which each skeleton $\tilde X$ is mapped to a common tangent space $T_{\tilde {X_0}}(C)$ at a reference shape $\tilde {X_0}$. The reference shape $\tilde {X_0}$ is defined as a pre-selected skeleton representing the neutral pose. Then, a transformation layer is built in this tangent space to increase global or local dissimilarities between class actions. This layer is followed by a CONV Block and a one-layer LSTM network, which learns the temporal dynamics of the sequences. As output, a fully connected block yields the corresponding action class.
The CONV block consists of two 1D convolution layers followed by a pooling layer.
For end-to-end network training, we use the cross-entropy loss as the training loss.


\subsection{Optimization over rigid transformations}
To optimize over rigid transformations, 3D rotations are applied to individual skeletons across sequences within this layer, and are updated during the training step. 
  
Let $\tilde{Y_i}$ denote the $i^{th}$ centered, unit norm skeleton in a sequence $S$, and $\hat{Y_i}$ its representative in the tangent space (reshaped from a $3(n-3)$ vector into a $3\times (n-1)$ matrix represented in the ambient coordinates). The transformation layer is performed on each sequence resulting in a hidden output $h$, given by:
\begin{equation}
    h_i = O_i\hat{Y_i} 
\end{equation}
where $O_i \in SO(3)$.
In the back-propagation phase, the gradient descent adapts the kernels $O_i$ directly so that they may not lie in $SO(3)$. To ensure that the updated kernels lie in $SO(3)$, we propose a second variant of this layer, denoted angle-based, where the optimization is performed over the rotation angles. Rotation matrices are then generated in the feed-forward pass.

\subsection{Optimization over non rigid transformation}

The optimization over local transformations is performed by finding the best rotations of 3D skeleton joints, with respect to the $x$, $y$ and $z$ axes, that improve performance on the action recognition task.  

Let $\tilde{Y_i}$ denote the $i^{th}$ centered, unit norm skeleton in a sequence $S$,  $\hat{Y_i}$ its representative in the tangent space (reshaped from a $3(n-3)$ vector into a $3\times (n-1)$ matrix represented in the ambient coordinates), and $q_i^j \in \mathbb{R}^3$ the $j^{th}$ joint of $\hat{Y_i}$. The transformation layer is performed on each sequence resulting in a hidden output $h$, given by:

\begin{equation}
    h_i = \{O_{i,j}q_i^j\}_{j=1}^n, \; \;
\end{equation}
where $O_{i,j} \in SO(3)$.

Similarly to the rigid transformation case, an angle-based optimization variant is proposed to ensure that each $O_{i,j}$ is a rotation matrix. 
In Section \ref{section:experiments}, we perform a study that compares the two variants for optimization over rigid and non rigid transformations: 1) the variant that allows the network to use general kernels as $3 \times 3$ matrices (not necessarily rotation matrices), and 2) the angle-based approach that constrains the network to allow rotation matrices only.

 

\section{Experimental results}
\label{section:experiments}

First, in Section \ref{section:Datasets and settings}, we first describe the datasets used to validate our architecture and experimental settings. For the demonstration of KShapeNet efficiency, an ablation study is presented in Section \ref{section:ablationstudy} with the discussion of the impact of intermediate layers, i.e., the transformation layer and logarithmic map layer. Then, in Section \ref{section:comparaisonStateOfArt}, we compare the performance of action recognition of the KShapeNet architecture to state-of-the-art approaches on the same datasets. We conclude in Section \ref{section:AdditionalStudies} with the comparison and discussion of the different variants of the transformation as well as the projection on tangent space layers. 

\subsection{Datasets and settings}
\label{section:Datasets and settings}

We evaluate the effectiveness of our KShapeNet framework on two large scale state-of-the-art datasets, NTU-RGB+D and NTU-RGB+D120. 
\\
\\
\textbf{NTU-RGB+D} \cite{Author42} is one of the largest 3D human
action recognition datasets. It consists of 56,000 action
clips of 60 classes. 40 participants have been asked to
perform these actions in a constrained lab environment,
with three camera views recorded simultaneously. The Kinect sensors estimate and record the 3D coordinates of 25 joints in the 3D camera’s coordinate system. For standard assessment, we utilize two state-of-the-art protocols: cross-subject (CS) and cross-view (CV). In the cross-subject protocol, the 40 subjects are split into training and testing sets (20 subjects each) made up of 40,320 and 16,560 samples, respectively. 
In the cross-view protocol, we select the samples from cameras 2 and 3 for training, and the samples from camera 1 for testing.
The training set then consists of the front and two side views of the actions, while the testing set incorporates left and right
45 degree views of the actions. For this assessment, the training and testing sets have 37,920 and 18,960 samples, respectively.\\
\\
\textbf{NTU120 RGB+D} (NTU120) \cite{Author43} is an extension of NTU60. It is the largest RGB+D dataset for 3D action recognition with 114,480 skeleton sequences. It contains 120 action classes performed by 106 distinct human subjects. For this dataset, the two protocols used for evaluation are cross-subject (CS) and cross-setup (Cset). For the cross-subject setting, half of the 106 subjects are used for training and the rest for testing. For the cross-setup setting, half of the setups are used for training and the rest for testing.

For the KShapeNet implementation, we set the number of frames to 100, and the batch size to 64 for the NTU dataset and 32 for the NTU 120 dataset. To estimate the model's parameters, we use the cross-entropy loss function and set the number of epochs to 30.  The Adam optimizer is adapted to train the network, and the initial learning rate is fixed to $1\times 10^{-4}$ for both datasets. For training the network, we used a machine with a processor speed of 3.40 GHz, memory of 32 GB and an NVIDIA GTX 1070 Ti GPU.
\subsection{Ablation study}
\label{section:ablationstudy}
In order to  validate the effectiveness of the proposed framework and highlight the impact of each processing block, we performed an ablation study by gradually adding 1) the projection to tangent space block, and 2) the transformation layer.
Table \ref{Ablation} reports the results of this study on the NTU and NTU120 datasets. In the first row, labeled "Baseline", we  illustrate the results of the deep learning network (CNN-LSTM used in KShapeNet) obtained with input data represented on the pre-shape space (without moving to the linear tangent space) and without the optimization over rigid or non rigid transformations. The baseline architecture actually presents fairly satisfactory results. However, they are not competitive to those produced by state-of-the-art approaches.

\begin{table}[ht]
\begin{center}
\begin{tabular}{|p{2cm}|p{1cm}|p{1cm}|p{1cm}|p{1cm}|}

\hline
Dataset &   \multicolumn{2}{|c|}{NTU-RGB+D} & \multicolumn{2}{|c|}{ NTU-RGB+D120} \\ \hline
\hline

 Protocol    &  CS          & CV     &  CS            & Cset \\ \hline
Baseline                         & 85.1              & 91.2         & 56.0         & 63.5                         \\ \hline
With transformation layer only                    & 89.6          & 91.5          &   57.2       & 63.8                                          \\ \hline
With projection to tangent space only           & 94.1           & 95.5    & 63.9          & 65.3                     \\ \hline
\textbf{Ours (KShapeNet)}        & \textbf{97.2}           & \textbf{96.2}        & \textbf{64.0}         & \textbf{66.1}                                                           \\ \hline
\end{tabular}
\end{center}
\caption{Ablation study results on the NTU and NTU120 datasets (\% accuracy).}
\label{Ablation} 
\end{table}

The second row of Table \ref{Ablation} depicts the results achieved by adding the transformation layer to the baseline architecture. The transformation layer adopted here considers optimization over non rigid transformations using the angle-based variant. Further discussion about the choice of this configuration is presented in Section \ref{section:TransformationVariants}. With reference to the "Baseline" results, the transformation layer improves the recognition performance by $4.5\%$ for CS and by $0.3\%$ for CV on the NTU dataset, and by $1.2\%$ for CS and $0.3\%$ for Cset on the NTU120. As explained in Section \ref{section:TransformationLayer}, this configuration of the transformation layer optimizes over non rigid transformations, hence urging the network to find the best local rotations that are applied to the the skeleton sequences; this justifies the improvement of action recognition accuracy. 

In the third row of Table \ref{Ablation}, we present the results obtained by only adding the projection to tangent space block to the baseline model. The tangent space projection provides significant improvements in recognition performance, jumping from $85.1\%$ to $94.1\%$ for the CS protocol on the NTU dataset. The increase in accuracy due to the projection of skeleton sequences to the tangent space is the result of a new skeleton representation in this Euclidean space, allowing for the definition of a linear distance metric between skeleton shapes.

In the fourth row of Table \ref{Ablation}, we report the final results produced by the KShapeNet framework, embedding both the projection on tangent space block and the transformation layer.
KShapeNet results in a significant improvement over the baseline model, and most importantly, further increases recognition accuracy over the two models with individually added components (the projection on tangent space block or transformation layer). It is worth to point out that the combination of both components empowers the network to properly discriminate action classes.
For instance, for the CS protocol on the NTU dataset, the accuracy increase due to the additional transformation layer was only $4.5\%$ and the increase due to the projection to tangent space block was only $9\%$. The addition of both the transformation layer and the projection to tangent space block (i.e., KShapeNet) increased recognition accuracy  by more than $12\%$. Accordingly, we conclude that the efficiency of KShapeNet is not only due to the advanced feature extraction capacity of the CNN-LSTM network, but equally due to the convenient data representation of skeleton shapes in the linear tangent space and the optimization over local rotational transformations.

\subsection{Comparison to state-of-the-art approaches}
\label{section:comparaisonStateOfArt}
In this section, we compare the performance of the proposed framework to state-of-the-art approaches on the two datasets, NTU and NTU 120. 

Table \ref{table:3} shows the results of the top performing state-of-the-art approaches on the NTU dataset, and compares them to the results of KShapeNet. 
In this table, we distinguish between three classes of action recognition methods: deep learning methods, Riemannian methods and hybrid (deep Riemannian) methods; our framework, KShapeNet, falls into the third category.  
The results demonstrate that KShapeNet consistently outperforms deep learning (leveraging CNNs and RNNs), Riemannian, and even hybrid approaches. Indeed, our method outperforms the best of these state-of-the-art approaches by $7.3\%$ and $0.1\%$ on the CS and CV settings, respectively. Comparing to the hybrid method of \cite{Author41}, in which the authors incorporate the Lie group structure into a deep network architecture using rotation mapping layers, our approach increases recognition accuracy by more than $35\%$. 

Table \ref{table:4} compares recognition accuracies between the most effective state-of-the-art approaches and KShapeNet on the NTU 120 dataset. In this case, KShapeNet achieves competitive recognition results for both the CS and Cset protocols. 

\begin{table*}[t]
\begin{center}
\begin{tabular}{ |p{6cm}|p{3cm}|p{3cm}|  }
 \hline
 \multicolumn{3}{|c|}{NTU-RGB+D Dataset} \\
 \hline
 \textbf{Deep learning methods} & Cross Subject &Cross View\\
 \hline
   \hline
 Directed Graph Neural Networks\cite{author49}   &89.9\%  &  96.1\%   \\
  \hline
Two stream adaptive GCN\cite{author50} &    88.5\%   & 95.1\% \\
  \hline
LSTM based RNN\cite{author51} & 89.2\% & 95.0\% \\
  \hline
AGC-LSTM(Joints\&Part)\cite{author52} &  89.2\% & 95.0\% \\
  \hline
  \hline
\textbf{Riemannian methods} & Cross Subject & Cross View\\
 \hline
 
Lie Group \cite{Author59}&    50.1\%  &  52.8\% \\ 
 \hline
Intrinsic SCDL \cite{Author58}   &    73.89\%  &  82.95\% \\ 
 \hline
  \hline
\textbf{Deep Riemannian methods} & Cross Subject & Cross View\\
\hline
Deep learning on $SO(3)^n$ \cite{Author41} &    61.37\%  &  66.95\% \\
 \hline
 \textbf{ Ours (KShapeNet)} &    \textbf{97.2\%}  &  \textbf{96.2\% }\\
 \hline 
\end{tabular}
\end{center}
\caption{Comparison to state-of-the-art top performing approaches on the NTU dataset.}
\label{table:3}
\end{table*}

\begin{table*}[t]
\begin{center}

\begin{tabular}{ |p{6cm}|p{3cm}|p{3cm}|  }
 \hline
 \multicolumn{3}{|c|}{NTU-RGB+D Dataset120} \\
 \hline
 \textbf{Method} & Cross Subject & Cross Setup\\
 \hline
 Tree Structure + CNN\cite{author53}   & 67.9\%  & 62.8\%\\
  \hline
 SkeleMotion\cite{author54}&   67.7\%  & 66.9\%   \\
  \hline
 Body Pose Evolution Map\cite{author55} &64.6\%  & 66.9\%\\
  \hline
Multi-Task CNN with RotClips\cite{Author56}    &62.2\%  & 61.8\%\\
  \hline
\textbf{Ours (KShapeNet)} &    \textbf{64.0\% } &  \textbf{66.1\%}\\
 \hline
\end{tabular}
\end{center}
\caption{Comparison to state-of-the-art top performing approaches on the NTU 120 dataset.}
\label{table:4}
\end{table*}

\subsection{Additional studies}
\label{section:AdditionalStudies}

Next, we present intermediate experiments that were performed during the design of KShapeNet. In particular, we discuss the different configurations that were tested in terms of the variants of the transformation layer and the projection onto tangent space block.

\subsubsection{Comparison of transformation layer variants} 
\label{section:TransformationVariants}

Table \ref{table:transformationLayer} presents a comparison of the four different variants of the transformation layer, based on the recognition results of KShapeNet, for the NTU and NTU120 datasets. Each row in Table \ref{table:transformationLayer} refers to one of the four tested settings of the transformation layer: 1) optimization over rigid transformations using the rotation matrix based variant (Rigid Matrix), 2) optimization over rigid transformations using the angle-based variant (Rigid Angle), 3) optimization over non rigid transformations using the rotation matrix based variant (NonRigid Matrix), and 4) optimization over non rigid transformations using the angle-based variant (NonRigid).

\begin{table}[ht]
\begin{center}
\begin{tabular}{|p{3cm}|p{0.6cm}|p{0.6cm}|p{0.6cm}|p{0.6cm}|}
\hline
Dataset &   \multicolumn{2}{|c|}{{\footnotesize NTU-RGB+D}} & \multicolumn{2}{|c|}{ {\footnotesize NTU-RGB+D120}} \\ \hline
Protocol    &   CS          & CV   &  CS           & Cset  \\ \hline
\hline
{\footnotesize Rigid Matrix based}               &  97.0 & 97.1           &64.4 & 65.7  \\ \hline
{\footnotesize Rigid Angle based}        & 96.9 &96.3         & 63.2 &  64.8     \\ \hline
{\footnotesize NonRigid Matrix based}       & 96.8       & 96.9         & 63.0 & 64.9  \\ \hline
{\footnotesize NonRigid Angle based}  & \textbf{97.2} & \textbf{96.2}    & \textbf{64.0} &   \textbf{66.1}    \\ \hline
\end{tabular}
\end{center}
\caption{Comparison of different variants of the transformation layer (\% accuracy).}
\label{table:transformationLayer} 
\end{table}

At a global level, we notice that for all of the transformation layer variants, the CS evaluation protocol results in similar performance to the CV protocol on the NTU dataset. On the other hand, the Cset evaluation protocol performs better than the CS protocol on the NTU 120 dataset. This second result suggests that the transformation layer is particularly important in discerning view variations in the Cset evaluation protocol (which includes different configurations of cameras capturing the activity). At a granular level, we highlight two different behaviors of the optimization over rigid transformations and the optimization over non rigid transformations, with regards to the two different variants: rotation matrix-based and angle-based. On the one hand, the rotation matrix-based variant, which gives the network the liberty to optimize matrix coefficients without any constraints (updated matrices may not be in $SO(3)$), yields better results for the optimization over rigid transformations than for the optimization over non rigid ones. On the other hand, the angle-based variant, which only updates the angles resulting in elements of $SO(3)$, performs worse for rigid transformations than non rigid ones.


Rigid transformations, i.e., rotations of the entire skeleton, are characterized by preserving the skeleton's shape, distance and angle properties (i.e., all joints move in the same direction by the same amount). We argue that, for this reason, the rotation matrix-based variant is more adequate for the optimization over such transformations. In other words, the rigid transformation is not subject to shape and angle variations, and the network tends to perceive the transformations applied to the skeleton as a one entity operation. Therefore, it is more efficient to allow the network to freely optimize over matrices during the back forward phase without the orthogonality constraint.
As a result of the non rigid transformations, i.e., different rotations applied to all of the joints, the shape and angle properties of the skeletons are not preserved at each pass. Beyond the first feed forward pass, the network will alter the representation of each sequence. Thus, for the optimization over non rigid transformations, it is more convenient to constrain the network to allow rotations only. The rotation matrices are generated based on updated rotation angles, always resulting in elements of $SO(3)$.


For the final configuration of KShapeNet, we chose to optimize over non rigid transformations using the angle-based variant; the corresponding recognition results are highlighted in bold in Table \ref{table:transformationLayer}. While not always resulting in best recognition accuracy, this choice allows for flexible modeling of inter-joint transformations.

\subsubsection{Comparison of projection to tangent space methods}
\label{section:ProjectionToTangentSpace}
As another intermediate experiment, we tested two variants of the projection to tangent space block. The first variant uses the logarithmic map to project all skeleton sequences to a single tangent space defined at a neutral reference skeleton. In this variant, the distances between skeleton shapes computed in the tangent space are different than those computed directly on Kendall's shape space, which introduces distortion (the only distances that are preserved after the projection are those from the reference to each projected shape). The issue is exacerbated when projecting skeleton shapes that are far away from the reference skeleton. Since all first frames of all skeleton sequences in the considered datasets are neutral, i.e., they are very close to each other on Kendall's shape space, we can alternatively map each sequence to the tangent space defined at the skeleton shape corresponding to its first frame; we again use the logarithmic map for this projection. 
The pitfalls of this second variant are twofold: 1) distance computations are no longer executed between points in the same Euclidean space, but between points in a set of "nearby" planes, and 2) the tangent spaces generally have different coordinate systems.

\begin{table}[ht]
\begin{center}
\begin{tabular}{|p{2cm}|p{1cm}|p{1cm}|p{1cm}|p{1cm}|}

\hline
Dataset &   \multicolumn{2}{|c|}{NTU-RGB+D} & \multicolumn{2}{|c|}{ NTU-RGB+D120} \\ \hline
\hline
 Protocol    &  CS          & CV   & CS            & CSet\\ \hline

{\footnotesize Log map (Same sequence frame) }   & 97.2              & 96.1          & 64.0         & 66.1               
   \\ \hline
{\footnotesize Log map ($1^{st}$ frame)} & 97.0              & 98.5        & 64.8       & 64.9                                      \\ \hline
\end{tabular}
\end{center}
\caption{Comparison of different variants of projection to tangent space using the logarithmic map (\% accuracy).}
\label{tab:Comapre_log} 
\end{table}

Table \ref{tab:Comapre_log} presents a comparison of recognition accuracies produced by the two variants on the NTU and NTU 120 datasets. Some of the results reported in this table suggest that the projection on the tangent space defined at the first frame of each sequence yields better performance over the projection on the tangent space defined at the same reference point for all sequences. However, this improvement is not consistent. Additionally, we expect that when the first frames across all sequences are not very similar, this result will not generalize. Thus, in KShapeNet, we opted to use the logarithmic map to project all sequences on the tangent space defined at the same neutral reference skeleton.

To push the capabilities of our model, we next tried to incorporate parallel transport (PT) (refer to Section \ref{section:Modeling of shape space trajectories}) as an alternative approach to map the skeleton sequences from the preshape space to the tangent space. In this approach, we first compute the shooting vectors between each consecutive frame within each sequence (using the logarithmic map). We then use PT to map these shooting vectors to the tangent space at the reference skeleton shape. Table \ref{LogvsPT} presents the results of applying the one-shot logarithmic map (same as row 1 in Table \ref{tab:Comapre_log}) and the PT approach on the NTU and NTU 12O datasets.

\begin{table}[t]
\begin{center}
\begin{tabular}{|p{2cm}|p{1cm}|p{1cm}|p{1cm}|p{1cm}|}

\hline
Dataset &   \multicolumn{2}{|c|}{NTU-RGB+D} & \multicolumn{2}{|c|}{ NTU-RGB+D120} \\ \hline
\hline
 Protocol    &  CS          & CV   & CS            & CSet\\ \hline

Log map                        & 97.2              & 96.1          & 64.0         & 66.1               
   \\ \hline
Parallel Transport      & 96.8              & 96.7         & 64.2          & 64.2                                    \\ \hline
\end{tabular}
\end{center}
\caption{Comparison of performance when projecting to tangent space at the same reference skeleton using the logarithmic map and when using parallel transport (\% accuracy).}
\label{LogvsPT} 
\end{table}

Theoretically, PT should perform better than the direct projection to a tangent space at the reference skeleton shape since it remedies the distortion issues mentioned earlier. Nevertheless, as shown in Table \ref{LogvsPT}, the simpler approach, paradoxically, tends to outperform the PT approach based on overall accuracy. In our implementation, the mapping to the tangent space iterations were not performed along the whole geodesic path, because this would have been computationally expensive. In fact, considering the computation-accuracy improvement trade-off, we decided that it was not worth to iterate the PT mapping along the entire geodesic path. This in part justifies the better performance of the simple logarithmic map to a common reference point over the more complicated PT approach. 

At the end of the various experiments and studies that we conducted, we decided to adopt the following configurations for KShapeNet: projection on the tangent space using the logarithmic map with same reference, and optimization over non rigid transformations using the angle-based variant.

\section{Conclusion}
\label{section:conclusion}
In this paper, we proposed a shape space deep architecture, KShapeNet, for action recognition based on modeling human actions on Kendall's shape space. As part of our framework, we introduced a novel transformation layer to increase global or local dissimilarities between different types of actions. In the transformation layer, we optimize over rigid and non rigid transformations. In addition, we explored the use of two optimization variants: 1) rotation matrix-based, and 2) angle-based. We showed that the first variant "rotation matrix-based" is better suited for optimizing rigid transformations, while the second variant "angle-based" is more efficient for optimizing non rigid transformations. Extensive experiments, conducted on two challenging large benchmark datasets for action recognition, demonstrate that the proposed framework, KShapeNet, is consistently better or competitive compared to state-of-the-art approaches in terms of recognition accuracy.

In the future, we will explore extensions of shape space modeling and the proposed transformation layers through reinforcement of the KShapeNet architecture with a deeper learning network. As a following step, the skeleton representation and optimization methods proposed in this paper will be used to develop an unsupervised system for action recognition, going beyond the data labeling hindrance.


{\small
\bibliographystyle{ieee_fullname}
\bibliography{egbib}
}

\end{document}